\documentclass[11pt,a4paper]{article}
\usepackage[hyperref]{emnlp2018}
\usepackage{times}
\usepackage{latexsym}
\usepackage{url}
\usepackage{relsize}
\usepackage{url}

\usepackage{graphicx}
\usepackage{amsmath}

\newcommand{\ignore}[1]{}

\aclfinalcopy

\title{Identifying Well-formed Natural Language Questions}
\author{Manaal Faruqui\hspace{6mm}Dipanjan Das\\
Google AI Language}

\begin{document}

\maketitle

\begin{abstract}
Understanding search queries is a hard problem as it involves dealing with ``word salad" text ubiquitously 
issued by users.
However, if a query resembles a well-formed question, a natural language processing pipeline is able to perform more accurate interpretation, thus reducing downstream compounding errors.
Hence, identifying whether or not a query is well formed can enhance query understanding.
Here, we introduce a new task of identifying a well-formed natural language question.
We construct and release a dataset of 25,100 publicly available questions classified into well-formed and non-wellformed categories and report an accuracy
of 70.7\% on the test set.
We also show that our classifier can be used to improve
the performance of neural sequence-to-sequence models for generating questions for reading comprehension.
\end{abstract}

\section{Introduction}
User issued search queries often do not follow formal grammatical structure, and require 
specialized language processing \cite{bergsma2007learning,barr-jones-regelson:2008:EMNLP,manshadi2009semantic,mishra2011unsupervised}. Traditional natural language processing (NLP) tools trained on formal text (e.g. treebanks) often have difficulty analyzing search queries; the lack of regularity in the structure of queries makes it difficult to
train models that can optimally process the query to extract information that
can help understand the user intent behind the query \cite{baeza2006intention}.

\ignore{
\begin{table}
\centering
\begin{tabular}{|p{7cm}|}
\hline
\textbf{Non-wellformed query:} scooby doo breed\\
\textbf{Well-formed question:} What is the breed of scooby doo?\\
\hline
\end{tabular}
\label{fig:canonic}
\caption{An example showing a general web-query and a well-formed question.}
\end{table}
}

One clear direction to improve query processing is to annotate a large number of queries with the desired annotation scheme.
However, such an annotation can be prohibitively expensive and models trained on such queries might suffer from
freshness issues, as the domain and nature of queries evolve frequently
\cite{markatos2001caching,Bawa:2003,roy12}.
Another direction is to obtain a paraphrase of the given query that is a grammatical natural language
question, and then analyze that paraphrase to extract the required information
%(cf. Table~\ref{fig:canonic}) 
\cite{nogueira-cho:2017:EMNLP2017,DBLP:journals/corr/BuckBCGHGW17}.
There are available tools and datasets, such as Quora question
paraphrases
%\footnote{\url{https://data.quora.com/First-Quora-Dataset-Release-Question-Pairs}}
and the Paralex dataset \cite{paralex}
-- for identifying query paraphrases \cite{wang2017bilateral,tomar2017neural},
but these datasets do not contain information about whether a query is a natural
language question or not. Identifying well-formed natural language
questions can also facilitate a more natural interaction between a user and a machine in
personal assistants or chatbots \cite{Yang:2014:ARQ,mostafazadeh:2016} or while 
recommending related queries in search-engines.

Identifying a well-formed question should be
easy by parsing with a grammar, such as the English resource grammar 
\cite{copestake2000open}, but such grammars are highly precise 
and fail to parse more than half of web queries.
Thus, in this paper we present a model to predict whether a given query is a well-formed natural language question. 
We construct and publicly release a dataset of 25,100 queries
annotated with the probability of being a well-formed natural language question
(\S\ref{sec:data}). We then train a feed-forward neural network classifier that uses 
the lexical and syntactic features extracted from the query on this data (\S\ref{sec:model}).
On a test set of 3,850 queries, we report an accuracy of 70.1\% on the binary classification task.
We also demonstrate that such a query well-formedness classifier can 
be used 
to improve the quality of a sequence-to-sequence question generation model \cite{du2017learning} by showing an improvement of 0.2 BLEU score in its performance (\S\ref{sec:extrinsic}). Our dataset ise available for download at \url{http://goo.gl/language/query-wellformedness}.

\begin{table*}[!th]
  \centering
  \begin{tabular}{|lcl|}
  \hline
    Query & Well-formed? & Reasoning\\
  \hline
  what is the breed of scooby doo? & 1 & Grammatical and an explicit question \\
  tell me what’s the breed of scooby doo? & 0 & A command but not a question \\
  headache evenings? & 0 & Ungrammatical and not a question \\
  what causes headaches during evenings & 1 & Grammatical and an explicit question \\
  what 12.5 as a fraction? & 0 & An explicit question but ungrammatical \\
  \hline
  \end{tabular}
  \caption{Examples of well-formed and non-wellformed queries according to
  the annotation guideline.}
  \label{tab:anno}
\end{table*}

\section{Well-formed Natural Language Question Classifier}
\label{sec:classifier}

In this section we describe the data annotation, and the models used for
question well-formedness classification.

\subsection{Dataset Construction}
\label{sec:data}
We use the Paralex corpus \cite{paralex} that contains pairs of noisy paraphrase questions. These questions were issued by users in WikiAnswers (a Question-Answer forum) and consist of both web-search query like constructs (\textit{``5 parts of chloroplast?''}) and well-formed questions (\textit{``What is the punishment for grand theft?''}), and thus is a good resource for constructing the question well-formedness dataset.
We select 25,100 queries from the unique list of queries extracted from the corpus such
that no two queries in the selected set are paraphrases. The
queries are then annotated into well-formed or non-wellformed questions.
We define a query to be a well-formed natural language question if it satisfies
the following:
\begin{enumerate}
  \itemsep-0.3em
  \item Query is grammatical.
  \item Query is an explicit question.
  \item Query does not contain spelling errors.
\end{enumerate}

Table~\ref{tab:anno} shows some examples that were shown to the annotators
to illustrate each of the above conditions. Every query was labeled by five
different crowdworkers with a binary label indicating whether a query is
well-formed or not. We average the ratings of the five annotators to
get the probability of a query being well-formed. 
Table~\ref{tab:labeledqueries} shows some queries with obtained human annotation.
Humans are pretty good at identifying an implicit query (\textit{``Population of
owls...''}) or a simple well-formed question (\textit{``What is released...''}), but
may miss out on subtle spelling mistakes like ``disscovered'' or disagree on
whether the determiner ``the'' is needed before the word ``genocide'' (\textit{``What countries have genocide happened in?''}).
Similar to other NLP tasks like entailment \cite{dagan2006pascal,bowman:2015}, 
paraphrasing \cite{meandev} etc. we rely on the wisdom of the
crowd to get such annotations in order to make the
data collection scalable and language-independent.

\begin{table}[!tb]
\centering
\begin{tabular}{|p{5.5cm}|r|}
\hline
Query ($q$) & $p_{wf}(q)$ \\
\hline
population of owls just in north america? & 0.0 \\
who disscoverd rihanna? & 0.2 \\
what countries have genocide happened in? & 0.6 \\
what is released when an ion is formed? & 1.0 \\
\hline
\end{tabular}
\label{tab:labeledqueries}
\caption{Examples of human annotations on query well-formedness.}
\end{table}

\begin{figure}[!tb]
    \centering
    \includegraphics[scale=0.4]{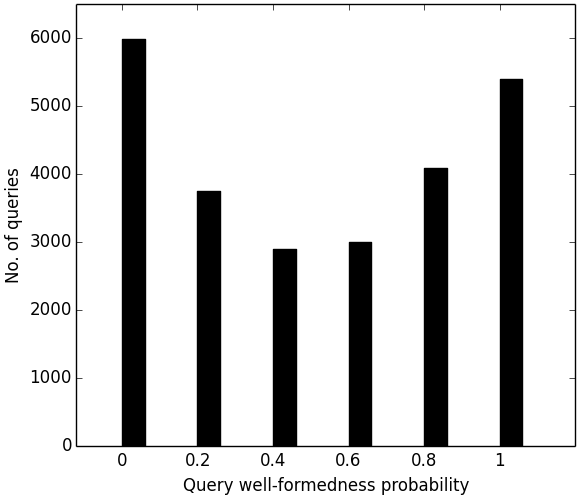}
    \caption{The distribution of the annotated questions according
    to well-formedness probability.}
    \label{fig:dist}
\end{figure}

Figure~\ref{fig:dist} is the histogram of query well-formedness probability across
the dataset. Interestingly, the number of queries where at least 4 or more annotators 
agree\footnote{At least 4 annotators label the query with 0 or 1.} on
well-formedness is large:
$|\{ q \mid 0.8 \le p_{wf}(q) \le 0.2\}| = 19206$
queries. These constitute $76.5\%$ of all queries in the dataset. 
%Following \newcite{meandev}, we compute a mean deviation 
%on the annotated dataset of $0.2$ which further shows strong annotator agreement.
The Fleiss' kappa \cite{fleiss1971measuring} for measuring agreement among multiple
annotators is computed to be $\kappa=0.52$ which shows moderate agreement \cite{landis1977measurement}.
We then randomly divided the dataset in approx. $70\%$, $15\%$, $15\%$ ratio into
training, development and test sets containing $17500$, $3750$, and $3850$ queries respectively.  While testing, we consider every query well-formed where at least 4 out of 5 
annotators ($p_{wf}\ge0.8 $) marked it as well-formed.\footnote{We randomly selected $100$ queries and
manually determined if each of those queries were well-formed. We found $p_{wf}(q)=0.8$ to be the
value above which all queries were well-formed.}

\subsection{Model}
\label{sec:model}

\begin{figure}[!tb]
    \centering
    \includegraphics[scale=0.5]{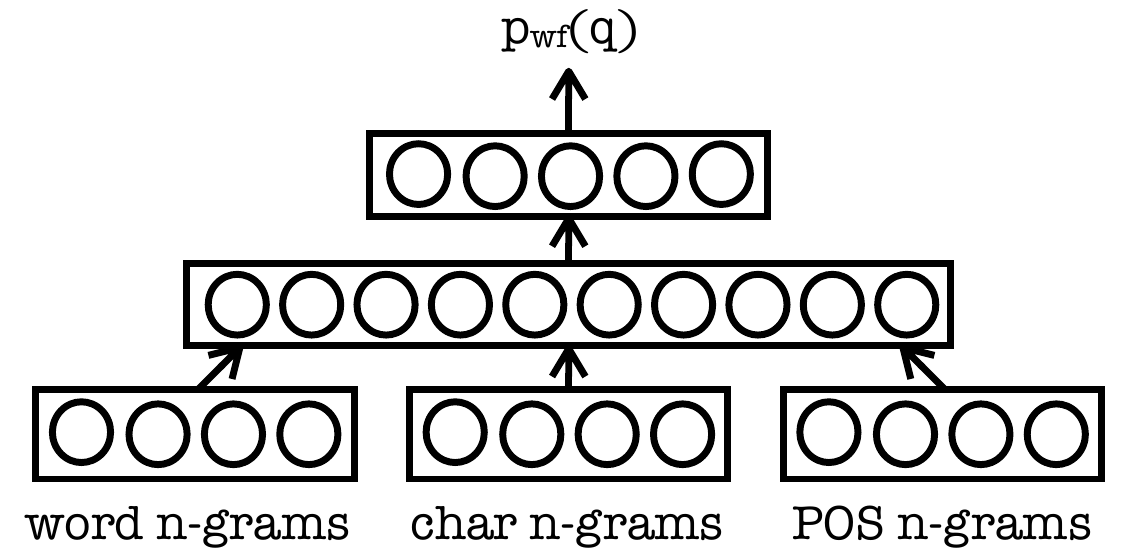}
    \caption{A feed-forward neural network for query well-formedness classification.}
    \label{fig:model}
\end{figure}

We use a feed-forward neural network with 2 hidden layers
with ReLU activations \cite{relu} on each layer and a softmax at the output layer
predicting $0$ or $1$. 
%The network architecture can be seen in Figure~\ref{fig:model}. 
We extract
a variety of features from the query which can be helpful in the classification. We
extract character-$3,4$-grams and word-$1,2$-grams as they can be helpful in capturing
spelling errors. In addition to lexical features, we also extract syntactic features that
can inform the model on any anomaly in the structure of the query. Specifically, we
annotate the query with POS-tags using SyntaxNet POS tagger
\cite{alberti-EtAl:2015:EMNLP} and extract POS-$1,2,3$-grams.\footnote{The use of dependency labels as features and use of pre-trained Glove embeddings
did not show improvement and hence omitted here for space constraints.}
Every feature in the network is represented as a real-valued embedding. All the $n$-grams
embeddings of every feature type are summed together and concatenated to form the input layer as shown in Figure~\ref{fig:model}.
The model is trained using cross-entropy loss against the gold labels for each
query. The hyperparameters are tuned to maximize accuracy
on the dev set and results are reported on the test set.% \mfar{Show accuracy on one more threshold.}

\paragraph{Hyperparameters.} 
We fix the size of the first and second hidden layers to be
$128$ and $64$ respectively. The character n-gram embeddings were of length $16$
and all other  feature embeddings were of length $25$. We use stochastic gradient
descent with momentum  for optimization with learning rate tuned over
$[0.001-0.3]$, a batch size of $32$ and $50000$ training steps.

\subsection{Experiments}

\begin{table}[!tb]
  \centering
  \begin{tabular}{|lr|}
  \hline
  Model & Accuracy ($\%$)\\
  \hline
  majority class baseline & $61.5$ \\
  % char bi-LSTM baseline & $64.4$ \\
  word bi-LSTM baseline & $65.8$ \\
  question word baseline & $54.9$\\
  \hline
  word-$1$ & $65.4$\\
  word-$1,2$ & $65.5$\\
  word-$1,2$ char-$3,4$ & $66.9$\\
  word-$1,2$ POS-$1,2,3$ & $\mathbf{70.7}$\\
  word-$1,2$ char-$3,4$ POS-$1,2,3$ & $70.2$\\
  \hline
  Approx. human upper bound & $88.4$\\
  \hline
  \end{tabular}
  \caption{Performance of well-formedness query classifier on the test set.}
  \label{tab:results}
\end{table}

\paragraph{Baselines.} The majority class baseline is $61.5\%$ which corresponds to all
queries being classified non-wellformed. The question word baseline that classifies any
query starting with a question word as a well-formed question gets
$54.9\%$.\footnote{List of question words: \url{https://en.wikipedia.org/wiki/Interrogative_word}} Also, we
used a single-layer word-level biLSTM encoder with hidden layer of length $50$
to encode the question and then use this representation in the softmax layer to
predict the label \cite{lee-dernoncourt:2016:N16-1}. This classifier achieved
$65.8\%$.

\paragraph{Results.} The best performance obtained is $70.7\%$ while
using word-$1,2$-grams and POS-$1,2,3$-grams as features. Using
POS $n$-grams gave a strong boost of $5.2$ points over word unigrams and bigrams.
Although character-$3,4$-grams gave improvement over word unigrams and bigrams,
the performance did not sustain when combined with POS tags.\footnote{We
assumed character $n$-grams to help identify spelling mistakes, but our dataset
has relatively few misspelled words--only 6 in 100 random queries.} A random sample of
1000 queries from the test set were annotated by one of the authors
of the paper with proficiency in English, which matched the gold label with
$88.4\%$ accuracy providing an approximate upper-bound for model performance.

A major source of error is our model's inability to understand deep semantics and syntax.
For example, \textit{``What is the history of dirk bikes?''} is labeled as a non-wellformed
question with $p_{wf}=0$ by annotators because of the misspelled word ``dirk'' (the correct 
word is ``dirt''). However, the POS tagger identifies ``dirk'' as a noun and as ``NN NNS'' 
is a frequent POS-bigram, our model tags it as a well-formed question with $p_{wf}=0.8$,
unable to identify that the word does not fit in the context of the question.
Another source of error is the inability to capture long term grammatical dependencies. For
example, in \textit{``What sort of work did Edvard Munch made ?''} the verb ``made'' is
incorrectly in the past tense instead of present tense. Our model is unable to capture
the relationship between ``did'' and ``made'' and thus marks this as a well-formed
question.

\section{Improving Question Generation}
\label{sec:extrinsic}

Automatic question generation is the task of generating questions that ask about the
information or facts present in either a given sentence or paragraph
\cite{Vanderwende:2008,Heilman:2010}. \newcite{du2017learning} present a state-of-the-art
neural sequence-to-sequence model to generate questions from a given sentence/paragraph.
The model used is an attention-based encoder-decoder network \cite{attention},
where the encoder reads in  a given text and the decoder is an LSTM RNN that
produces the  question by predicting one word at a time. \ignore{conditioned on the current
hidden layer and the attention-weighted hidden layers of the encoder.}

\newcite{du2017learning} use the SQuAD question-answering dataset \cite{squad} to
develop a question generation dataset by pairing sentences from the text with the
corresponding questions. 
The question generation dataset contains approx $70$k, $10$k, and $12$k training, development and test examples.
Their current best model selects the
top ranked question from the $n$-best list produced by the decoder as the output.
We augment their system by training a discriminative reranker
\cite{collins:2005} with the model score of the question generation model and the
well-formedness probability of our classifier as features to optimize BLEU score
\cite{bleu} between the selected question from the $10$-best list and the reference
question on the development set. We then use this reranker to select the best question
from the $10$-best list of the test set.

\begin{table}[!tb]
  \centering
  \begin{tabular}{|lrr|}
  \hline
  Model & BLEU-1 & BLEU-4\\
  \hline
  Baseline & $41.3$ & $12.0$ \\
  Reranked & $\mathbf{41.6}$ & $\mathbf{12.2}$ \\
  \hline
  \end{tabular}
  \caption{Reranking the $n$-best output of a neural seq2seq question generation model using well-formedness probability.}
  \label{tab:rerank}
\end{table}

We use the evaluation package released by \newcite{Chen2015MicrosoftCC} to compute
BLEU-1 and BLEU-4 scores.\footnote{BLEU-x uses precision computed over $\lbrack1, x\rbrack$-grams.} Table~\ref{tab:rerank} shows that the reranked question
selected using our query well-formedness classifier improves the BLEU-4 score of a
seq-to-seq question generation model from $12.0$ to $12.2$.
\ignore{Out of the $12$k questions in the test set, our model helped select a better
question  $449$ times which corresponds to $3.75\%$ of the total test set.}
The oracle improvement, by selecting the sentence from the list that maximizes
the BLEU-4 score is $15.2$.
However, its worth noting that an increase in well-formedness doesn't guarantee an improved
BLEU score, as the oracle sentence maximizing the BLEU score might be fairly non-wellformed
\cite{badbleu}. For example, \textit{``who was elected the president of notre dame in?''}
has a higher BLEU score to the reference \textit{``who was the president of notre dame in 1934?''}
than our well-formed question \textit{``who was elected the president of notre dame?''}.
Figure~\ref{fig:rerankexample}
shows a question generation example with the output of \newcite{du2017learning} 
as the baseline result and the reranked question using the wellformed probability.
%It can be seen that the reranked question being more grammatical is given a higher score of
%well-formedness by our classifier and results in a closer match to the gold question.

\begin{figure}[!tb]
\centering
%\small
\begin{tabular}{|p{7cm}|}
\hline
\textbf{Sentence:} montana is home to the rocky mountain elk foundation and has a historic big game hunting tradition.\\
\textbf{Gold question:} what is the name of the big game hunting foundation in montana?\\
\hline
\textbf{seq2seq:} what is a historic big game hunting tradition? ($p_{wf}=0.7$)\\
\textbf{Reranked:} what is the name of the historic big game tradition? ($p_{wf}=0.8$)\\
\hline
\end{tabular}
\label{fig:rerankexample}
\caption{Example showing question selection from the $n$-best list using our reranking model.}
\end{figure}

\section{Related Work}
We have referenced much of the related work throughout the paper. We now
review another orthogonally related field of work.
Grammatical error correction (GEC) is the task of correcting the grammatical
errors (if any) in a piece of text \cite{ng2014conll}. As GEC includes not
just identification of ungrammatical text but also correcting the text to
produce grammatical text, its a more complex task. However, grammatical error
prediction \cite{schmaltz-EtAl:2016:BEA11,daudaravicius-EtAl:2016:BEA11} is the
task of classifying whether or not a sentence is grammatical, which is more closely
related to our task as classifying a question as well-formed requires making
judgement on both the style and grammar of the text.

\section{Conclusion}
We proposed a new task of well-formed natural language question identification and established a strong baseline on a new dataset that can be downloaded at: \url{http://goo.gl/language/query-wellformedness}.
We also showed that question well-formedness information can be
a helpful signal in improving state-of-the-art question generation systems.

\bibliography{references}
\bibliographystyle{acl_natbib}

\end{document}